\documentclass[journal]{IEEEtranTIE}
\usepackage{graphicx}
\usepackage{cite}
\usepackage{picinpar}
\usepackage{amsmath}
\usepackage{url}
\usepackage{flushend}
\usepackage[latin1]{inputenc}
\usepackage{colortbl}
\usepackage{soul}
\usepackage{multirow}
\usepackage{pifont}
\usepackage{color}
\usepackage{alltt}
\usepackage[draft]{hyperref}
\usepackage{enumerate}
\usepackage{siunitx}
\usepackage{breakurl}
\usepackage{epstopdf}
\usepackage{pbox}
\usepackage{algorithm}
\usepackage{algorithmic}
\usepackage{stfloats}
\usepackage{multicol}
\usepackage{float}
\usepackage[switch] {lineno} 
\usepackage{amsmath}

\begin{document}

	\title{A Novel Dynamic Light-Section 3D Reconstruction Method for Wide-Range Sensing }
	
	\author{
		\vskip 1em
	Mengjuan Chen, Qing Li, Kohei Shimasaki, \emph{Member, IEEE}, Shaopeng Hu, Qingyi Gu, \emph{Member, IEEE}, Idaku Ishii, \emph{Member, IEEE}		\vskip 1.8em
		
		\thanks{
				Mengjuan Chen, Qing Li, Kohei Shimasaki, Shaopeng Hu, Idaku Ishii are with the Hiroshima University, Higashihiroshima 739-8527, Japan (e-mail: chenmengjuan2016@ia.ac.cn; soleilor-li@hiroshima-u.ac.jp;  shimasaki@robotics.hiroshima-u.ac.jp; hu@robotics.hiroshima-u.ac.jp;  iishii@robotics.hiroshima-u.ac.jp).
			
			Qingyi Gu is affiliated with the Research Center of Precision Sensing and Control, Institute of Automation, Chinese Academy of Sciences, Beijing 100190, China (e-mail: qingyi.gu@ia.ac.cn).
		}
	}
	\maketitle
	
	\begin{abstract}
		
		Existing galvanometer-based laser scanning systems are challenging to apply in multi-scale 3D reconstruction because of the difficulty in achieving a balance between high reconstruction accuracy and a wide reconstruction range. This paper presents a novel method that synchronizes laser scanning by switching the field-of-view (FOV) of a camera using multi-galvanometers. In addition to the advanced hardware setup, we establish a comprehensive mathematical model of the system by modeling dynamic camera, dynamic laser, and their combined interaction. We then propose a high-precision and flexible calibration method by constructing an error model and minimizing the objective function. Finally, we evaluate the performance of the proposed system by scanning standard components. The evaluation results demonstrate that the accuracy of the proposed 3D reconstruction system achieves 0.3 mm when the measurement range is extended to 1100 mm $\times$ 1300 mm $\times$ 650 mm. With the same reconstruction accuracy, the reconstruction range is expanded by a factor of 25, indicating that the proposed method simultaneously allows for high-precision and wide-range 3D reconstruction in industrial applications.
	
	\end{abstract}

%\def\abstractname{Note to Practitioners}
%\begin{abstract}
%	This article was motivated by the multi-scale 3D reconstruction challenge in laser scanning systems for industrial applications. Existing approaches can not simultaneously balance reconstruction range and accuracy, limiting their ability to meet the high-precision 3D reconstruction demands of large-scale targets in industrial settings. In this article, we propose a novel approach that utilizes multi-galvanometer systems to synchronize the control of laser and camera scanning, effectively overcoming the trade-off between accuracy and range. We mathematically model the proposed system and present a comprehensive high-precision calibration method. Experiments demonstrate that our system achieves high-precision and large-range performance. Compared to all existing galvanometer-based laser scanning methods, our approach offers the highest measurement range while maintaining the same level of measurement accuracy. The proposed system not only expands the reconstruction range but also allows for dynamic adjustment of reconstruction positions. In future research, we will focus on this application.
%\end{abstract}

	\begin{IEEEkeywords}
		 Dynamic 3D Reconstruction, Multi-Galvanometers, Light-Section, Calibration.
	\end{IEEEkeywords}
	
	\markboth{IEEE TRANSACTIONS ON Instrumentation and Measurement}%
	{}
	
	\definecolor{limegreen}{rgb}{0.2, 0.8, 0.2}
	\definecolor{forestgreen}{rgb}{0.13, 0.55, 0.13}
	\definecolor{greenhtml}{rgb}{0.0, 0.5, 0.0}
	
	\section{Introduction}
	
	\IEEEPARstart{L}{ight} -section vision systems are widely used in many applications for their
	adaptability, high accuracy, and effective cost \cite{article49,article35,article36}, such as rail traffic monitoring \cite{article34}, medical imaging \cite{article7}, robotics \cite{article10}, and industrial production\cite{article48}. Such systems typically comprise a camera, laser projector, and mechanical scanning platform. The line laser projects laser stripes onto the surface of the object, whereas the camera captures an image of the object with the laser stripes. The three-dimensional (3D) geometric information of the object is then obtained by triangulation, as extensively reviewed in literature \cite{article6}. The 3D reconstruction of an object can be completed by passing laser stripes or objects through a mechanical scanning platform.

	Traditional laser scanners rely primarily on mechanical driver shafts, which are large, complex, and slow \cite{article15}, \cite{article16}. To overcome these limitations, various scanning mechanisms have been proposed. For instance, Du \cite{article17} designed a system that mounts a line laser on the end of a robotic arm to improve scanning flexibility; however, its scanning accuracy was limited by the precision of the robotic arm. Jiang \cite{article18} proposed a system that uses gimbals to drive the laser and camera for scanning; however, the system size was significant, and its scanning speed was slow. In recent years, galvanometers have emerged as promising scanning devices because of their small size, fast rotation, and high control accuracy. This galvanometer-based solution provides a better alternative in terms of laser scanning accuracy and speed \cite{article41}. However, existing galvanometer-based laser-scanning systems are primarily designed to perform laser scanning while leaving the camera fixed. The limited FOV of a fixed camera causes a trade-off between the accuracy and the sensing range of the system, which significantly affects its efficiency. 

	In this study, we propose a novel dynamic light-section 3D reconstruction system that combines dynamic laser and dynamic camera using multi-galvanometers. Our approach utilizes multiple galvanometers to synchronize laser scanning and the FOV switching of the camera, thereby enabling high-precision and wide-range 3D reconstruction. Calibration is required to achieve this, which includes the system calibration of the galvanometer-based dynamic laser and camera, and their joint calibration. 
	For calibrating galvanometer-based dynamic laser systems, Eisert \cite{article19} introduced a mathematical model and calibration procedure; however, the model was complicated, and its optimization was difficult, thus leading to low accuracy. Yu \cite{article20} designed a one-mirror galvanometer laser scanner. However, the calibration procedure was complex, and the objective function was difficult to optimize. Similarly, Yang \cite{article21} proposed a calibration method based on a precision linear stage. However, this approach relies on a precision instrument and lacks flexibility. 
	For calibrating galvanometer-based dynamic camera systems, Ying \emph{et al.}, \cite{article22,article23,article24} introduced self-calibration methods, which were complex in theory and difficult to implement. Kumar \cite{article25} proposed a calibration method based on the look-up table (LUT) using simple linear parameters, which required complex pre-processing. Junejo \emph{et al.}, \cite{article27,article28,article29,article30} proposed feature-based calibration methods, which were time-consuming and had low accuracy. Han\cite{article38} introduced a calibration method for galvanometer-based camera using an end-to-end single-hidden layer feed forward neural network model, but it was computationally intensive. Boi \emph{et al.}, \cite{article46} \cite{article47} proposed manifold constrained Gaussian process regression methods for galvanometer setup calibration, which relied on data-driven and complex calibration procedures. Hu \cite{article31} built a galvanometer mirror-based stereo vision measurement system and established a mirror reflection model, but it still lacked an accurate calibration method.
	
	In conclusion, current light-section 3D reconstruction systems cannot simultaneously have high accuracy and wide range. Moreover, existing calibration methods only focus on calibrating either dynamic lasers or dynamic cameras  and still have some shortcomings, as mentioned above. To address these limitations, this study proposes a novel dynamic 3D reconstruction system that overcomes the trade-off between accuracy and measurement range by synchronizing laser scanning and FOV switching of a camera based on multiple galvanometers. Additionally, we propose a complete calibration solution for the proposed system that includes the calibration of the dynamic camera, dynamic laser, and joint calibration. The contributions of this study can be summarized as follows:

\begin{itemize}
\setlength{\leftskip}{-0.75em}
\item[1)]
    A novel dynamic light-section 3D reconstruction system is designed based on multiple galvanometers. To the best of our knowledge, the system is the first to synchronize laser scanning and FOV switching of camera, thus enabling high-precision and wide-range 3D reconstruction simultaneously.
\end{itemize}

	\begin{itemize}
	\setlength{\leftskip}{-0.75em}
	\item[2)]
	The geometrical model of the dynamic 3D system is developed by the combined modeling of the dynamic camera and laser, which establishes 3D-information acquisition with multi-galvanometers and laser images.
    \end{itemize}

	\begin{itemize}
	\setlength{\leftskip}{-0.75em}
	\item[3)]
	A flexible and accurate calibration method of the dynamic 3D system is proposed by constructing error models and objective functions. This method is not only applicable to the proposed system  but also to other single galvanometer-based laser or camera systems.

    \end{itemize}

	\begin{itemize}
	\setlength{\leftskip}{-0.75em}
	\item[4)]
	
	Experiments are conducted to validate the proposed dynamic 3D reconstruction method and demonstrate its accuracy. To the best of our knowledge, compared to all existing galvanometer-based laser scanning methods, our approach has the highest measurement range while maintaining the same level of measurement accuracy.
	\end{itemize}

    The system design and geometric model are described in Section II. The proposed calibration method and error compensation methods are described in Section III. Section IV presents the validation experiments and results. Finally, Section V presents the conclusions.

	\section{System Design and Geometric Model}
	\subsection{System Design}

\begin{figure}[H]
		\centering
		\includegraphics[width=0.35\textwidth]{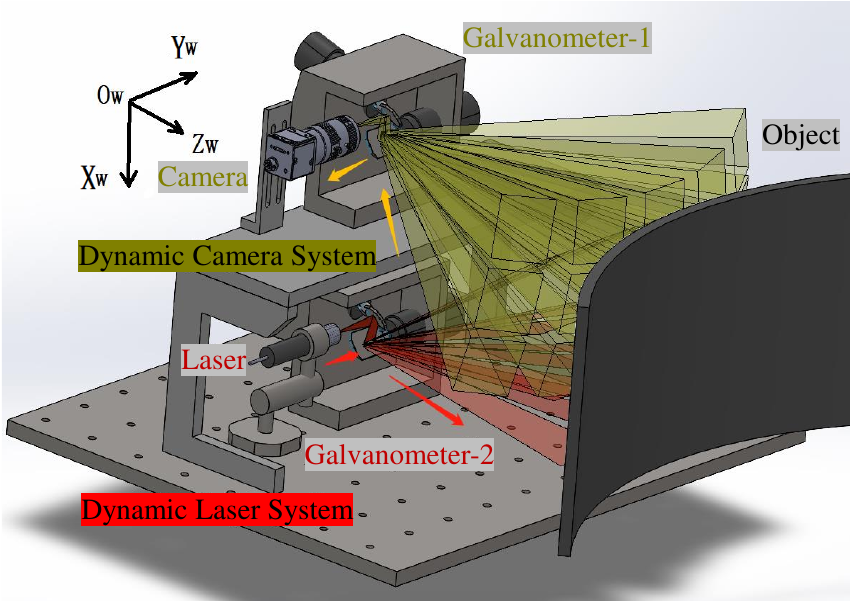}
		\caption{System design.}
		\label{Fig1}
	\end{figure}

	The dynamic light-section 3D reconstruction system consists of a CMOS camera, a line laser, and two galvanometer mirror systems, as shown in Fig. \ref{Fig1}. The camera and Galvanometer-1 form a dynamic camera system, whereas the laser and Galvanometer-2 form a dynamic laser system. Based on the mathematical model of the system and pre-calibration, the 3D information of the target can be calculated from the captured laser image and voltage values of the two galvanometers.

	\begin{figure*}[htb]
	\centering
	\includegraphics[width=0.95\textwidth]{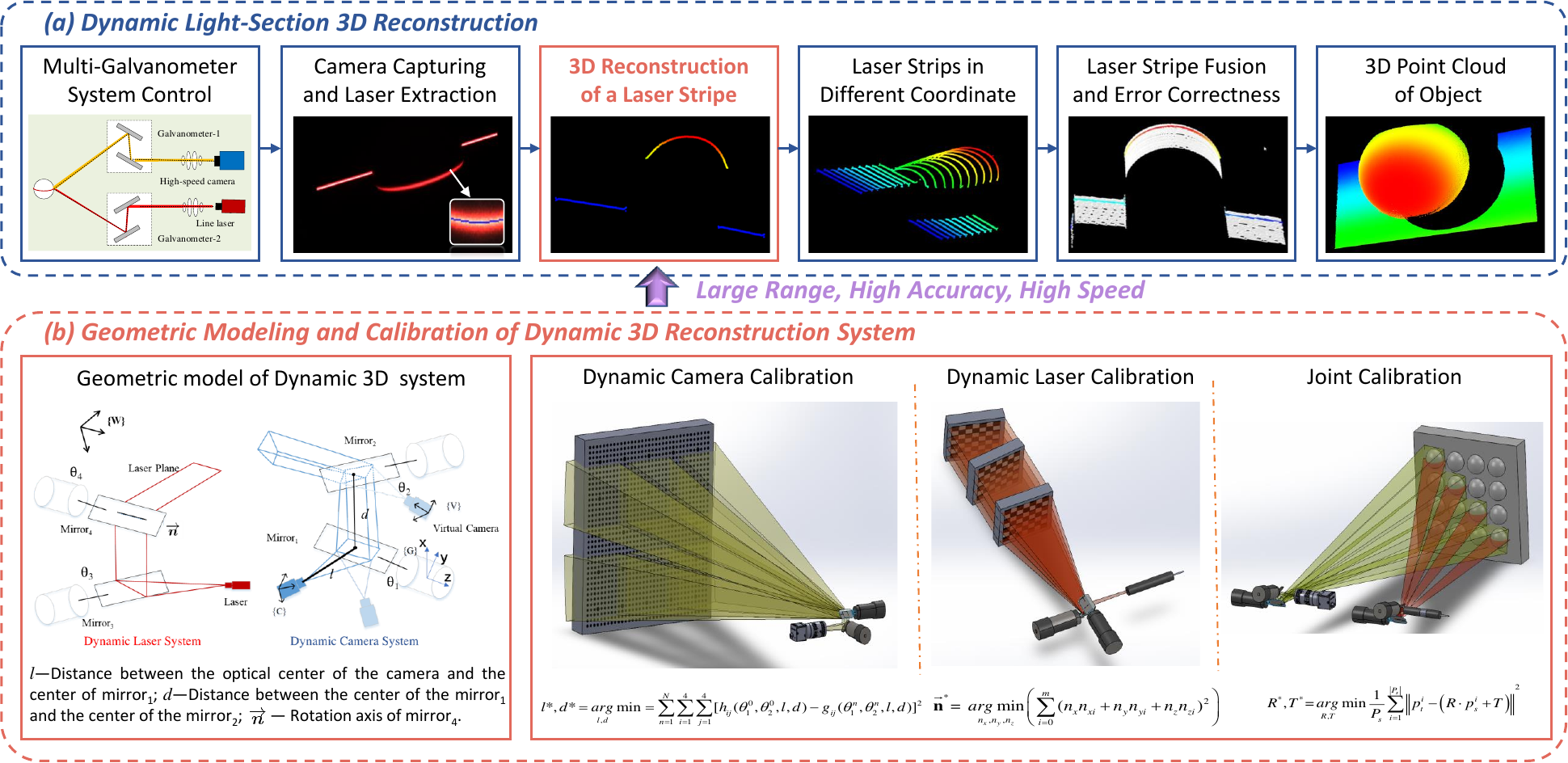}
	\caption{Framework of dynamic 3D reconstruction method. (a) The flowchart of dynamic light-section 3D reconstruction. (b) Geometric modeling and calibration of dynamic 3D reconstruction system. }
	\label{Fig2}
\end{figure*}
	
%\begin{figure}[H]
%	\centering
%	\includegraphics[width=0.35\textwidth]{images/FIG3_crop.pdf}
%	\caption{Geometric model of dynamic 3D system.}
%	\label{Fig3}
%\end{figure}

 The working principle is illustrated in Fig. \ref{Fig2}. A spherical object on a flat plane is employed to demonstrate the process of dynamic 3D reconstruction. First, the system utilizes multi-galvanometer control to scan the target surface. When the system is activated, a line laser projects a laser stripe onto Galvanometer-2, which reflects the stripe onto the surface of the object. By controlling the voltage of Galvanometer-2, the laser stripe can scan the target. Simultaneously, the dynamic camera system captures laser images from different angles by adjusting the voltage of Galvanometer-1. Next, the laser-center-pixel coordinates are obtained using the laser stripe extraction algorithm.
 The 3D reconstruction of the laser stripe is performed by combining the voltage values of multiple galvanometers, laser pixel coordinates, geometric models of the dynamic 3D reconstruction system, and calibrated parameters.
 The point clouds of all the laser stripes are converted to the same coordinate frame using the transfer matrix of the dynamic camera.
 The system performs error correction based on the joint calibration to optimize accuracy.
 Finally, the system generates a point cloud for the target and completes the dynamic 3D reconstruction. Accurate mathematical modeling and calibration methods are essential to ensure the 3D reconstruction accuracy of the system.

\subsection{Geometric Model}
	
	The mathematical model of the system is shown in Fig. \ref{Fig2}(b). Five coordinate frames are created: image pixel coordinate frame $\{I\}$, camera coordinate frame $\{C\}$, virtual camera coordinate frame $\{V\}$, world coordinate frame $\{W\}$, and Galvanometer-1 mirror coordinate frame $\{G\}$. According to the operating principle of a galvanometer, the rotation angle of the pan-tilt mirror is proportional to the voltage. For Galvanometer-1, the voltages of the pan-tilt mirrors are denoted by $U_{1-pan}$ and $ U_{1-tilt} $. Therefore, the rotation angle of the pan mirror is $\theta_1=k_{1-pan}U_{1-pan}$ and the rotation angle of the tilt mirror is $\theta_2=k_{1-tilt}U_{1-tilt}$. As the pan-tilt mirror rotates, $\{V\}$ reflects the change in $U_{1-pan}$ and $U_{1-tilt}$. 
	When $U_{1-pan} = U_{1-tilt} = 0$, the virtual camera coordinate frame is denoted by $\{V_0\}$. The relationship between the virtual coordinate frame $\{V\}$ after the motion and the initial coordinate frame $\{V_0\}$ is given by Eq. (\ref{equ-1}).

	\begin{align}
		\label{equ-1}
		\{V\} = {^V}\boldsymbol{T}_{V_0}{\{V_0\}} = {^V}\boldsymbol{T}_{G}{^G}\boldsymbol{T}_{V_0}\{{V_0\}},
	\end{align}
where ${^V}\boldsymbol{T}_{V_0}$ is the transfer matrix between ${\{V_0\}}$ and ${\{V\}}$. ${^V}\boldsymbol{T}_{G}$ denotes the transfer matrix between ${\{G\}}$ and $\{{V}\}$. ${^G}\boldsymbol{T}_{V_0}$ denotes the transfer matrix between ${\{V_0\}}$ and ${\{G\}}$. As shown in the geometric model diagram, ${\{C\}}$ is first reflected by a pan mirror and then by a tilt mirror. The geometries of the two reflections are modeled using Eq. (\ref{equ-2}). For ${\{V_0\}}$, the rotation angle of the pan-tilt mirror is $\theta_1 = \theta_2 = 45^{\circ}$. The transfer matrix ${^V}\boldsymbol{T}_{V_0}$ is calculated using Eq. (\ref{equ-3}). Thus, the geometric model of the dynamic camera is established.

\begin{figure*}[!t]
	\begin{equation}
		\label{equ-2}
		\begin{split}
			\fontsize{8pt}{8pt}
			{^V}\boldsymbol{T}_G = 
			\begin{bmatrix}
				1 & 0 & 0 & 0 \\
				0 & \cos 2\theta_2 & \sin2 \theta_2 & d(1-\cos 2\theta_2)\\
				0 & \sin 2\theta_2 & -\cos 2\theta_2 & -d\sin 2\theta_2\\
				0 & 0 & 0 & 1
			\end{bmatrix}
			\begin{bmatrix}
				-\cos2\theta_1 & \sin2\theta_1 & 0 & 0 \\
				\sin2\theta_1 & \cos2\theta_1 & 0 & 0 \\
				0 & 0 & 1 & 0 \\
				0 & 0 & 0 & 1
			\end{bmatrix}
			\begin{bmatrix}
				0 & 0 & 1 & -l \\
				1 & 0 & 0 & 0 \\
				0 & 1 & 0 & 0 \\
				0 & 0 & 0 & 1
			\end{bmatrix} =\\
			\fontsize{8pt}{8pt}
			\begin{bmatrix}			
				\sin2\theta_1 & 0 & -\cos2\theta_1 & l\cos2\theta_1 \\
				\cos2\theta_1\cos2\theta_2 & \sin2\theta_2 & \sin2\theta_1\sin2\theta_2 & -l\sin2\theta_1\cos2\theta_2+d(1-\cos2\theta_2) \\
				\cos2\theta_1\sin2\theta_2 & -\cos2\theta_2 & \sin2\theta_1\sin2\theta_2 & -l\sin2\theta_1\sin2\theta_2-d\sin2\theta_2 \\
				0 & 0 & 0 & 1
			\end{bmatrix}.
		\end{split}	    
	\end{equation}
\end{figure*}

\begin{equation}
	\label{equ-3}
	{^V}\boldsymbol{T}_{V_0}={^V}\boldsymbol{T}_G{{^{V_0}}\boldsymbol{T}_G}^{-1}|(U_{1-pan}=U_{1-tilt}=0).
\end{equation}

The coordinates of the pixel points on the laser stripe are denoted by $(u, v)$. The coordinates of the corresponding 3D points in $\{V\}$ are denoted by $(X_V, Y_V, Z_V)$. Based on the pinhole model of the camera, the mapping relationship between their coordinates can be obtained using Eq. (\ref{equ-5}).

%\begin{equation}
%	\label{equ-5}
%	\begin{bmatrix}
%		u \\
%		v \\
%		1
%	\end{bmatrix}
%	=
%	\begin{bmatrix}
%		f_x & 0 & u_0 & 0\\
%		0 & f_y & v_0 & 0\\
%		0 & 0 & 1 & 0
%	\end{bmatrix}
%	\begin{bmatrix}
%		\boldsymbol{R} & \boldsymbol{T}\\
%		0^T & 1
%	\end{bmatrix}
%	\begin{bmatrix}
%		X_V \\
%		Y_V \\
%		Z_V \\
%		1
%	\end{bmatrix}.
%\end{equation}

\begin{equation}
	\label{equ-5}
	Z_V\left[\begin{array}{l}
		u \\
		v \\
		1
	\end{array}\right]=\left[\begin{array}{ccc}
		f_x & 0 & u_0 \\
		0 & f_y & v_0 \\
		0 & 0 & 1
	\end{array}\right]\left[\begin{array}{c}
		X_V \\
		Y_V \\
		Z_V
	\end{array}\right],
\end{equation}
where $(u_0, v_0)$ is the principal point of the image, $f_x$ and $f_y$ are the focal length of the camera. For Galvanometer-2, the rotation angle of the pan mirror is denoted by $\theta_3=k_{2-pan}U_{2-pan}$, and the rotation angle of the tilt mirror is denoted by $\theta_4 = k_{2-tilt}U_{2-tilt}$. When $U_{2-pan} = U_{2-tilt} = 0$, the dynamic laser is in its initial position. The initial laser plane in $\{V_0\}$ is denoted by $^{V_0}plane_0$ and its equation is  $A_{0}x + B_{0}y + C_{0}z + D_{0} = 0$. The rotation axis of the dynamic laser in $\{V_0\}$ is denoted as $\overrightarrow{\boldsymbol{n}}=(n_x,n_y,n_z)$. The plane of the laser after rotation about the rotation axis is denoted as $^{V_0}plane$, and the equation is $A_{V_0}x + B_{V_0}y + C_{V_0}z + D_{V_0} = 0$. 

The light path of the dynamic laser is reflected by a mirror and rotated along its axis. The rotation angle of the laser plane is twice that of the mirror plane. Therefore, the equation for the dynamic laser plane after rotation in $\{V_0\}$ can be solved using Eq. (\ref{equ-6}),

\begin{equation}
	\label{equ-6}
	\left[\begin{array}{c}A_{V_0} \\ B_{V_0} \\ C_{V_0}\end{array}\right]=\boldsymbol{R}\left(\overrightarrow{\mathbf{n}}, 2 \theta_4\right)\left[\begin{array}{c}A_0 \\ B_0 \\ C_0\end{array}\right],
\end{equation}
where $\boldsymbol{R}$ represents the Rodrigues transformation. Using a point $(x_n,y_n,z_n)$ on the rotation axis, $D_{V_0}$ can be calculated as $D_{V0} = -A_{V_0}x_n - B_{V_0}y_n - C_{V_0}z_n$. Combining this with the transfer matrix ${^V}\boldsymbol{T}_{V0}$ in Eq. (\ref{equ-3}), the equation for the dynamic laser plane in $\{V\}$ can be calculated as

\begin{equation}
	\label{equ-7}
	{{^V}}plane = {{^V}}\boldsymbol{T}_{{V_0}}{^{{V_0}}}plane.
\end{equation}

The equation for ${^V}plane$ is denoted as $A_Vx+B_Vy+C_Vz+D_V=0$. By extracting the pixel points $(u,v)$ from the laser stripe, the corresponding 3D point can be calculated as ${^V}\boldsymbol{P} = (X_V, Y_V, Z_V)$. Therefore, the relationship between ${^V}\boldsymbol{P}$ and the change in the galvanometer mirror angles can be expressed by Eq. (\ref{equ-8}).

\begin{equation}
	\label{equ-8}
	\fontsize{9pt}{10pt}
	\left\{  
	\begin{array}{l}
		Z_V=D_V/[A_V\times(u-u_0)/f_x+B_V\times(v-v_0)/f_y+C_V] \\ 
		X_V=(u-u_0)/f_x\times Z_V \\ 
		Y_V=(v-v_0)/f_y\times Z_V \\ 
		\left\{A_V, B_V, C_V, D_V\right\}=\boldsymbol{F}\left(\theta_1, \theta_2, \theta_4 ; A_0, B_0, C_0, D_0, l, d, \overrightarrow{\mathbf{n}}, \mathbf{P}\right).    
	\end{array}
	\right.
\end{equation}

$\boldsymbol{F}$ represents a parameterized mapping function, where $\theta_1, \theta_2, \theta_4$ are variables and other parameters are constants. Because $\{V\}$ changes constantly with the scanning angle, it is necessary to convert all ${^V}\boldsymbol{P}$ into a coordinate frame that is fixed with $\{W\}$. For ease of calculation, we choose $\{V_0\}$ and obtain ${^{V_0}}\boldsymbol{P}= (X_{V_0}, Y_{V_0}, Z_{V_0})= {^{V_0}}\boldsymbol{T}_V{^V}\boldsymbol{P}$.

Thus, we establish the relationship between $(u, v)$ and $(X_{V_0}, Y_{V_0}, Z_{V_0})$ to formulate the 3D reconstruction. In the mathematical model of the 3D dynamic system, Eq. (\ref{equ-8}) shows that $f_x,f_y,u_0,v_0$ can be obtained by calibrating the camera. $A_0, B_0, C_0, D_0$ can be obtained from the laser plane calibration. $ l, d, \mathbf{\overrightarrow{\boldsymbol{n}}} = (n_x, n_y, n_z), \boldsymbol{P} = (X_0, Y_0, Z_0)$ are unknown, and a calibration algorithm must be designed to obtain the unknowns.

\section{Calibration Method}

The proposed system calibration method is divided into three parts: the dynamic camera calibration, the dynamic laser calibration, and the joint calibration of the dynamic camera and laser for error correction.

\subsection{Dynamic Camera Calibration}
To calibrate the intrinsic parameters, $f_x, f_y, u_0, v_0$, of the camera, we use the method proposed by Zhang\cite{article32}. The dynamic camera calibration method described in Section II is used to obtain the constraint relationship between $\{V\}$ and $\{G\}$, as shown  in Eq. (\ref{equ-2}); thus, we obtain the parameters $l$ and $d$.

The proposed calibration method uses a large calibration board as shown in Fig. \ref{Fig2}(b). The calibration board measures 740 $\times$ 740 mm and comprises a total of 35 $\times$ 35 circular markers. These markers are constructed from 25 individual 7 $\times$ 7 sub-patterns. Each sub-pattern features a central larger circular marker with a diameter of 15 mm, while the remaining smaller circular markers have a diameter of 10 mm, with a center-to-center spacing of 20 mm. The purpose of the larger circular markers is to establish the relationships between calibration points across different FOVs. The calibration board is scanned by varying the galvanometer voltage to obtain numerous images at different rotation angles. Each image corresponds to a virtual coordinate frame. The number of images is denoted as $n$. The conversion matrix between $\{V\}$ and $\{W\}$ can be obtained by extrinsic parameter calibration as ${^{V_0}}\boldsymbol{T}_W, {^{V_1}}\boldsymbol{T}_W, {^{V_2}}\boldsymbol{T}_W,\ldots, {^{V_n}}\boldsymbol{T}_W$. As the relative positions between the calibration points in these images are known, the transformation matrix between virtual coordinate frames is calculated as ${^{V_1}}\boldsymbol{T}_{V_0}, {^{V_2}}\boldsymbol{T}_{V_0},\ldots,{^{V_n}}\boldsymbol{T}_{V_0}$. These values are taken as observations. Multiple sets of observations are used to solve for the parameters to be calibrated. The initial pan-tilt angles of Galvanometer-1 are denoted by $\theta_{1}^{(0)}$ and $\theta_{2}^{(0)}$. They are the corresponding angles of $\{V_0\}$ and the first calibration image. The pan-tilt angles of Galvanometer-1 corresponding to $\{V_n\}$ and $n$th calibration images are denoted as $\theta_1^{(n)}$ and $\theta_2^{(n)}$. Therefore, $\{V_0\}$ and $\{V_n\}$ are defined as follows:

\begin{equation}
	\label{equ-9-2}
	\begin{array}{c}
		\{V_0\} = k_{ij}(\theta_1^{(0)}, \theta_2^{(0)}, l, d), \\
		\{V_n\} = g_{ij}(\theta_1^{(n)}, \theta_2^{(n)}, l, d),
	\end{array}
\end{equation}
where $k_{ij}$ and $g_{ij}$ represent the function of $\{V_0\}$ and $\{V_n\}$. $l$ and $d$ are parameters to be calibrated using Eq. (\ref{equ-8}). The transformation matrix between $\{V_0\}$ and $\{V_n\}$ is a 4$\times$4 matrix, which can be expressed as:

\begin{equation}
	\label{equ-10}
	{ }^{V_{n}} \mathbf{T}_{V_{0}}(n)=\left[\begin{array}{cccc}
		a_{11}^{n} & a_{12}^{n} & a_{13}^{n} & a_{14}^{n} \\
		a_{21}^{n} & a_{22}^{n} & a_{23}^{n} & a_{24}^{n} \\
		a_{31}^{n} & a_{32}^{n} & a_{33}^{n} & a_{34}^{n} \\
		0 & 0 & 0 & 1
	\end{array}\right]=a_{i j}^{n}.
\end{equation}

Simultaneously, $\{V_n\}$ can be calculated using $\{V_0\}$ from Eq. (\ref{equ-9-2}), and $^{V_n}\boldsymbol{T}_{V_0}$. For ease of representation, this is denoted as $h_{ij}$.

\begin{equation}
	\begin{aligned}
		\begin{split}
			\label{equ-11}
			&\left\{{V}_{n}\right\}={ }^{V_n} \boldsymbol{T}_{V_{0}}(n) \cdot\left\{{V}_{0}\right\}=\\
			&\sum_{i=1}^{4} \sum_{j=1}^{4} a_{i j}^{n} k_{i j}\left(\theta_{1}^{(0)}, \theta_{2}^{(0)}, l, d\right)
			=h_{i j}\left(\theta_{1}^{(0)}, \theta_{2}^{(0)}, l, d\right).
		\end{split}
	\end{aligned}
\end{equation}

$\{G\}$ is the base coordinate frame and $\{V\} = ^V\boldsymbol{T}_G \cdot \{G\}$; therefore, Eq.~(\ref{equ-2}) is the mathematical model of $\{V\}$. Eq. (\ref{equ-11}) is the result of $\{V\}$ obtained from multiple observations. Eq. (\ref{equ-9-2}) shows the results calculated using the mathematical model of the dynamic camera. For all the measured coordinate frames ($\{V_0\}, \{V_1\}, \{V_2\},\ldots,\{V_n\}$), the sum of the errors between the theoretical and measured values must be minimized. Therefore, the objective function is formulated using Eq. (\ref{equ-13}).

\begin{equation}
	\begin{split}
		\label{equ-13}
		l *, d *=\underset{l, d}{\arg \min } \sum_{n=1}^{N} \sum_{i=1}^{4} \sum_{j=1}^{4}\left[h_{i j}\left(\theta_{1}^{0}, \theta_{2}^{0}, l, d\right)-g_{i j}\left(\theta_{1}^{n}, \theta_{2}^{n}, l, d\right)\right]^{2}.
	\end{split}
\end{equation}

In Eq. (\ref{equ-2}), the parameters $l$ and $d$ exist only in translation vector. Based on the objective function, Eq. (\ref{equ-16}) can be obtained. Finally, the parameters $l$ and $d$ are obtained by solving Eq. (\ref{equ-16}) using the least-square method.

\begin{equation}
	\label{equ-16}
	\fontsize{9pt}{10pt}
	\left\{\begin{array}{l}
		g_{14}\left(\theta_{1}^{(1)}, \theta_{2}^{(1)}, l, d\right)-h_{14}\left(\theta_{1}^{(0)}, \theta_{2}^{(0)}, l, d\right)=0 \\
		g_{24}\left(\theta_{1}^{(1)}, \theta_{2}^{(1)}, l, d\right)-h_{24}\left(\theta_{1}^{(0)}, \theta_{2}^{(0)}, l, d\right)=0 \\
		g_{24}\left(\theta_{1}^{(1)}, \theta_{2}^{(1)}, l, d\right)-h_{24}\left(\theta_{1}^{(0)}, \theta_{2}^{(0)}, l, d\right)=0 \\
		\ldots \\
		g_{14}\left(\theta_{1}^{(n)}, \theta_{2}^{(n)}, l, d\right)-h_{14}\left(\theta_{1}^{(0)}, \theta_{2}^{(0)}, l, d\right)=0 \\
		g_{24}\left(\theta_{1}^{(n)}, \theta_{2}^{(n)}, l, d\right)-h_{24}\left(\theta_{1}^{(0)}, \theta_{2}^{(0)}, l, d\right)=0 \\
		g_{24}\left(\theta_{1}^{(n)}, \theta_{2}^{(n)}, l, d\right)-h_{24}\left(\theta_{1}^{(0)}, \theta_{2}^{(0)}, l, d\right)=0.
	\end{array}\right.
\end{equation}

\subsection{Dynamic Laser Calibration}
Based on the mathematical modeling of the dynamic laser presented in Section II, the equation of $^{V_0}plane_0$ and the rotation axis of the dynamic laser $\overrightarrow{\boldsymbol{n}}$ must be calibrated when Galvanometer-1 is at the initial position. The calibration of $^{V_0}plane_0$ is conducted utilizing the methodology outlined in \cite{article33}. This process involve the acquisition of laser images at various positions using a checkerboard calibration plate. Through this approach, the laser plane is accurately defined by fitting multiple laser lines. For the extraction of the laser center, the technique described in \cite{article50} is employed, offering the advantage of sub-pixel precision in the extraction process. Following these procedures, we successfully derive the equation $A_{0}x + B_{0}y + C_{0}z + D_0 = 0$.

The Galvanometer-2 voltage $U_{2-tilt}=U_1, U_2,\ldots, U_m$ are used to move the laser and obtain multiple laser planes. Next, the respective equations are calibrated in the same manner as $^{V_0}plane_0$ and denoted as $^{V_0}plane_1, ^{V_0}plane_2,\ldots,^{V_0}plane_m$. The unit normal vectors of these planes are calculated as $\overrightarrow{\boldsymbol{n}}_0 (n_{x_0}, n_{y_0}, n_{z_0})$, $\overrightarrow{\boldsymbol{n}}_1(n_{x_1}, n_{y_1}, n_{z_1})$, $\overrightarrow{\boldsymbol{n}}_2(n_{x_2}, n_{y_2}, n_{z_2}),\ldots, \overrightarrow{\boldsymbol{n}}_m(n_{x_m}, n_{y_m}, n_{z_m})$. In the absence of errors, the laser planes intersect along the same straight line. This line is the laser rotation axis $\overrightarrow{\boldsymbol{n}}$ and is also the rotation axis of the tilt mirror in Galvanometer-2. For a normal vector in any laser plane, we obtain $\overrightarrow{\boldsymbol{n}} \cdot \overrightarrow{\boldsymbol{n}}_i = 0 (i = 0, 1, 2,\ldots,m)$. However, $\overrightarrow{\boldsymbol{n}} \cdot \overrightarrow{\boldsymbol{n}}_i$ do not exactly equal zero because of various errors. Therefore, for all laser planes, the objective function is formulated as shown in Eq. (\ref{equ-17-1}).

\begin{equation}
	\begin{split}
		\overrightarrow{\mathbf{n}}^{*}=
		\underset{n_{x}, n_{y}, n_{z}}{\arg \min }\left(\sum_{i=0}^{m}\left(n_{x} n_{x_ i}+n_{y} n_{y_i}+n_{z} n_{z_i}\right)^{2}\right)
		\label{equ-17-1}.
	\end{split}
\end{equation}

The direction vector $\overrightarrow{\boldsymbol{n}}$ of the rotation axis is obtained by minimizing the objective function. $\boldsymbol{P} = (X_0, Y_0, Z_0)$ is a point on the rotation axis in all laser planes and can be obtained using the least-square method. All parameters in Eq.~(\ref{equ-8}) are obtained by calibrating the camera, laser plane, dynamic camera, and dynamic laser. Thus, we complete the calibration of the proposed dynamic 3D system.

\subsection{Joint Calibration for Error Correction}

For a well-calibrated dynamic light-section 3D reconstruction system, there are two sources of error, dynamic camera and dynamic laser, as listed in Table \ref{tab-3}.

This study proposes an error correction method based on the joint calibration of a dynamic camera and dynamic laser. After the calibration is completed, error correction is performed based on the 3D reconstructed results. Theoretically, when Galvanometer-1 is scanning and Galvanometer-2 is fixed, the reconstructed laser point cloud coincides perfectly. However, as explained in the error source analysis, there is some deviation between the multiple laser point clouds owing to these errors. We correct these errors using point-cloud registration to obtain the accuracy conversion matrix.

The calibration process is designed based on the following principle. If Galvanometer-2 remains stationary, the laser will be out of the FOV after Galvanometer-1 has scanned a certain range. Therefore, multiple calibration positions must be set in advance to maintain the laser in the FOV. These are set in advance as $p_1,p_2,\ldots,p_n$ and the corresponding voltages of Galvanometer-2 at these positions are ${(s_1, t_1), (s_2, t_2),\ldots,(s_n, t_n)}$, respectively. The laser plane equations {$plane_{p_1}, plane_{p_2},\ldots,plane_{p_n}$} at these positions are calibrated using the laser plane calibration method described in Part B. The voltage of Galvanometer-1 is denoted by ($s_1^\prime$, $t_1^\prime$), ($s_2^\prime$, $t_2^\prime$)$,\ldots,$($s_m^\prime$, $t_m^\prime$), respectively. The error-correction flow is presented in Algorithm \ref{alg3}.

\begin{algorithm} 
	\caption{Joint Calibration for Error Correction} 
	\label{alg3} 
	\begin{algorithmic}[1]
		\STATE \textbf{Input:} ${f_x, f_y, u_0, v_0}$;${l, d}$;${A_1, B_1, C_1, D_1,\ldots,A_n, B_n, C_n,}$\\${ D_n}$;${(s_1, t_1), (s_2, t_2),\ldots,(s_n, t_n)}$.
		\STATE \textbf{Output:} $^{V_0}\boldsymbol{T}_{V_1},^{V_0}\boldsymbol{T}_{V_2},\ldots,^{V_0}\boldsymbol{T}_{V_m}$.		
		\STATE\textbf{Initialize:} {$i$ $\gets$ 1, $(s, t)$ $\gets$ $(s_1, t_1)$, $(s^\prime,t^\prime)$ $\gets$ $f(s,t)$, du $\gets$ 0.1, $^{V_1}plane$ $\gets$ $plane_{p_1}$, $\boldsymbol{M}$ $\gets$ Unit matrix.}
		\STATE{Using Eq. (\ref{equ-8}) to obtain the laser point cloud $^{V_0}\boldsymbol{P}_1$}.
		\WHILE{$i < m$}
		\STATE{Laser Image capture and center curve extraction};
		\STATE{$^{V_i}plane = g(s,t,s^\prime,t^\prime)$};
		\STATE{Using Eqs. (\ref{equ-1})(\ref{equ-2})(\ref{equ-3}) to calculate $^V \boldsymbol{T}_{V_i}$};
		\STATE{ $^Vplane $ $\gets$ $ ^V \boldsymbol{T}_{V_i} {}^{V_i}plane$};
		\STATE{Using Eq. (\ref{equ-8}) to obtain the point cloud $^{V_0} \boldsymbol{P}_i$}.
		\STATE{ $^V \boldsymbol{T}_{V_i}$ $\gets $ Transfer matrix between $^{V_0}\boldsymbol{P}_1$ and $^{V_0}\boldsymbol{P}_i$};
		\STATE{$^{V_0}\boldsymbol{T}_{V_i}$ $\gets$ $ \boldsymbol{M} ^V\boldsymbol{T}_{V_i} $};
		\IF{$i \% 50 == 0$}
		\STATE{$(s, t)$ $\gets$ $(s_{(i/50)}, t_{(i/50)})$, $^{V_i}plane \gets ^{V}plane$};
		\STATE{Using Step. (8) $-$ Step. (11) to obtain point cloud $\boldsymbol{P}$}.
		\STATE{$\boldsymbol{M}$ $\gets $ The transfer matrix between $\boldsymbol{P}$ and $^{V_0}\boldsymbol{P}_i$};
		
		\ENDIF
		
		\STATE{$i \gets i + 1$};
		\STATE{$(s^\prime,t^\prime)$ $\gets (s^\prime + du \times i \% 200, t^\prime + 1 \times \ {int} (i / 200) )$};
		\ENDWHILE
		
	\end{algorithmic} 
\end{algorithm}

\section{Experiment}

\begin{figure}[H]
	\centering
	\includegraphics[width=0.22\textwidth]{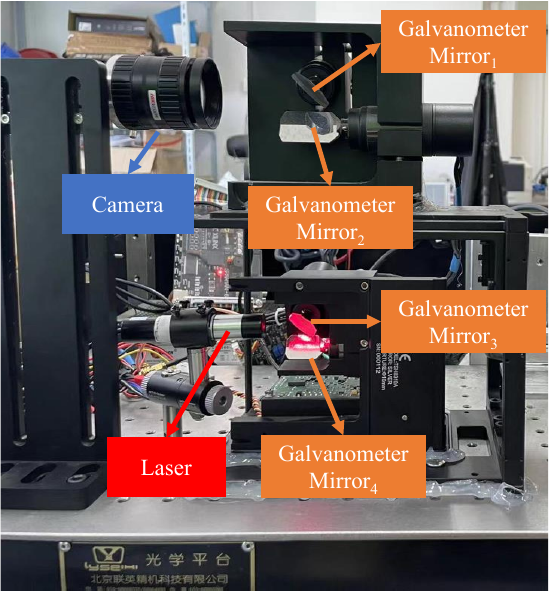}
	\caption{3D dynamic reconstruction system based on multiple galvanometers and light section.}
	\label{Fig5}
\end{figure}

The proposed dynamic 3D reconstruction system is built as shown in Fig. \ref{Fig5}. The camera model is MV-CA004-10UC, with a pixel size of 6.9 $\mu$m $\times$ 6.9 $\mu$m, resolution of 720 pixels $\times$ 540 pixels, and frame rate of 500 fps. The exposure time of the camera to capture the dark image of the laser is 500 $\mu$s. The laser model is LXL65050-16 and the laser wavelength is 650 nm.  

The models of both Galvanometer-1 and Galvanometer-2 are TSH8310. The galvanometer is used to scan a range of  $\pm20^{\circ}$ using a control voltage range from $- 10 V$ to $+ 10 V$. The maximum scan frequency is 1 kHz, with an angular resolution of 0.0008, thus, the system has the potential for high accuracy and resolution.

\subsection{Calibration Accuracy Verification}

\subsubsection{Dynamic camera calibration accuracy}
Twenty-five images of the calibration board are collected when $U_{1-pan}=U_{1-tilt}=0$. The camera is calibrated using Zhang's \cite{article32} calibration method in OpenCV. After calibration, the intrinsic parameters are $f_x=$ 7801.38, $f_y = $ 7798.24, $u_0 = $ 359.51, and $v_0 = $ 269.54. The focal length is 53.83 mm. According to the calibration method for the dynamic cameras presented in Section III, the system parameters are solved as $l$ = 83.45 mm and $d$ = 22.14 mm.

Based on these calibration results, the mathematical model proposed in Section II can be used to calculate the theoretical transfer matrix for the pan-tilt mirror of Galvanometer-1 at different angles. The transfer matrices corresponding to these angles are directly measured using a calibration board. The matrix 2-norm is calculated according to Eq. (\ref{equ-18-1}) to compare the theoretical matrices $\boldsymbol{A}$ and measured transfer matrices $\boldsymbol{B}$ for calibration accuracy verification. The pan-tilt voltages of Galvanometer-1 are varied from $- 10V$ to $10V$, at intervals of $4V$. Thirty-six positions are measured. The error between the theoretical and measured transfer matrices is obtained, and the error curves are shown in Fig. \ref{Fig6}(a). The results show that the RMSE (Root Mean Square Error) is 1.231 mm between the theoretical and measured values.

\begin{equation}
	\label{equ-18-1}
	E(\boldsymbol{A}, \boldsymbol{B})=\sqrt{\sum_{i=1}^4 \sum_{j=1}^4\left(\boldsymbol{A}_{i j}-\boldsymbol{B}_{i j}\right)^2}
\end{equation}

This confirms the accuracy of the dynamic camera calibration. The observed errors originate from the geometric model and the calibration process, as explained in the error analysis section. It is important to note that the measured values obtained for the virtual camera using the calibration board may also exhibit slight deviations. 

Consequently, these findings serve as a validation of the accuracy of the dynamic camera calibration; however, these cannot be solely relied upon to assess the accuracy of the calibration. A more detailed accuracy verification can be conducted based on the outcomes of the 3D reconstruction analysis.

\subsubsection{Dynamic laser calibration accuracy}

\begin{figure*}[!htb]
	\centering
	\includegraphics[width=0.77\textwidth]{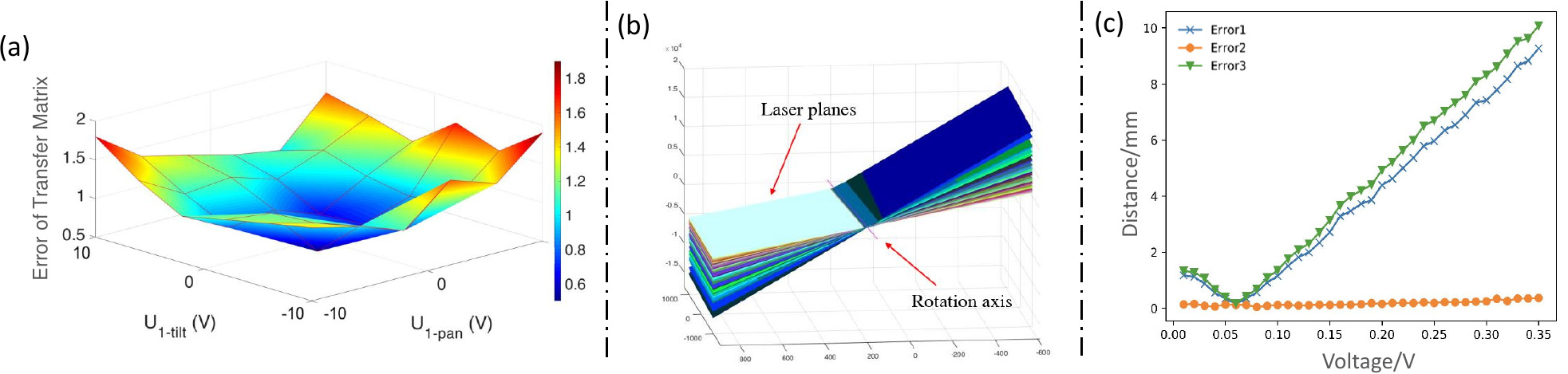}
	\caption{Calibration accuracy verification. (a) Error of dynamic camera transfer matrix. (b) Visualization of laser planes and the calibrated rotation axis. (c) Error curve before and after correction.}
	\label{Fig6}
\end{figure*}

With Galvanometer-1 fixed, the calibration board is positioned within the FOV of the virtual camera. The tilt mirror of Galvanometer-2 is rotated 30 times with a step size of $0.1V$, allowing the system to scan the calibration board, whose position is randomly changed five times (ensuring clear imaging in the virtual camera); the same 30 scans are repeated for each position. The laser rotation axis is solved as $(\overrightarrow{\mathbf{n}}, \mathbf{P})$ = $([0.99,0.02,-0.0004], [-18310.30, -195.93, 257.97])$. Fig. \ref{Fig6}(b) visually represents the laser plane and the rotation axis. Notably, the calibrated rotation axis align with the intersection of the laser planes, providing evidence for the accuracy of the dynamic laser calibration. A detailed accuracy assessment is subsequently performed by analyzing the results of the 3D reconstruction.

\begin{table*}
	\caption{Error Source and Analysis } 
	\label{tab-3} 
	\begin{center}
		\begin{tabular}{ |c|c|c| } 
			\hline	
			~ & Error & Source and Analysis \\ 
			\hline
			1& Rotation angle of galvanometer mirror $(\theta1 , \theta2 )$ & Non-linear deviations exists between voltage and rotation angle of Galvanometer-1. \\ 
			\hline
			2& Dynamic camera geometric model & The spectacular reflection geometric model has deviations with mechanical structure. \\
			\hline
			3& Calibration of parameter $(l,d)$ & This error has been optimized by proposed error model and objective function in this paper. \\
			\hline
			4& Camera intrinsic parameters calibration & This non-linear error is optimized using the Zhang's\cite{article32} calibration method. \\
			
			\hline
			5 & Rotation angle of galvanometer mirror $(\theta3 , \theta4 )$ & Non-linear deviations exists between voltage and rotation angle of Galvanometer-2. \\
			\hline
			6 & Dynamic laser geometric model & This error depends on the accuracy of the laser mechanical installation. \\
			\hline
			7 & Calibration of laser rotation axis & This error is optimized by the proposed objective function. \\
			\hline
			8 & Laser center curve extraction & Center extraction algorithm is based on the Hessian Matrix and ensures a high extraction accuracy. \\
			\hline
		\end{tabular}
	\end{center}
\end{table*}

\subsubsection{Joint calibration accuracy}
A calibration sphere is selected as the 3D reconstruction target for error correction. Galvanometer-2 is controlled to project the laser stripe onto the sphere, while Galvanometer-1 is fixed.The virtual camera, controlled by Galvanometer-1, captures images of the laser stripe from different views. The 3D reconstruction of these laser stripe images is performed based on the calibration results and mathematical models of the 3D dynamic system. The reconstructed point clouds, which are indicated as white, are shown in Fig. \ref{Fig2}(a), 'Error Correction'. Notably, white point clouds exhibit non-overlapping regions owing to errors. The correction method described in Algorithm \ref{alg3} is employed to register these white point clouds. The registration results are shown as colored point clouds in Fig. \ref{Fig2}(a). The distance between the point clouds before and after the correction is calculated to evaluate the error. The calculation formula is as follows:

\begin{equation}
	\label{equ-18}
	d=\frac{1}{P_{s}} \sum_{i=1}^{\left|P_{s}\right|}\left\|p_{t}^{i}-p_{s}^{i}\right\|^{2},
\end{equation}

Here, $p_s$ represents the point cloud of the laser stripe captured in the first virtual camera view, and $p_t$ represents the point cloud of the laser stripe captured from another view. The error between $p_t$ and $p_s$ is determined by performing a nearest-neighbor search, denoted as Error1. After the point-cloud registration, the error between $p_t$ and $p_s$ is calculated as Error2. In addition, the error before correction is computed as Error3 using the matched points from the point-cloud registration result. The error curves are shown in Fig. \ref{Fig6}(c). The RMSE of Error1 and Error3 before correction are calculated as 4.928 and 5.475 mm, respectively. However, after error correction, the RMSE of Error2 is significantly reduced to 0.197 mm. These results evidently indicate a substantial improvement in the accuracy following the error-correction process.

\subsection{3D Reconstruction Accuracy Verification}

\subsubsection{Standard blocks reconstruction test}

A standard stair block is employed to test the stability of the system and analyze its reconstruction accuracy at different angles. The stair block has a distance of 30 mm between its two planes, with machining errors within 1 $\mu$m. The 3D dynamic system proposed in the paper is used to reconstruct a stair block. Scanning is performed by synchronously controlling the tilt mirrors of both Galvanometer-1 and Galvanometer-2, rotating each by 0.1°.

Once the scanning and reconstruction processes are completed, a point cloud of the stair block is generated. Two planes (Plane-1 and Plane-2) of the stairs are fitted, and the distance between them is calculated. Point clouds belonging to Plane-2 are used to fit a plane equation using the least-square method. Next, 500 points belonging to Plane-1 are randomly selected and the average distance between these points and Plane-2 is calculated as the distance of the fitted plane. The difference between the calculated and actual distances is considered as the error, which serves as a measure of the reconstruction accuracy achieved by the system.

\begin{figure*}[!htb]
	\centering
	\includegraphics[width=0.95\textwidth]{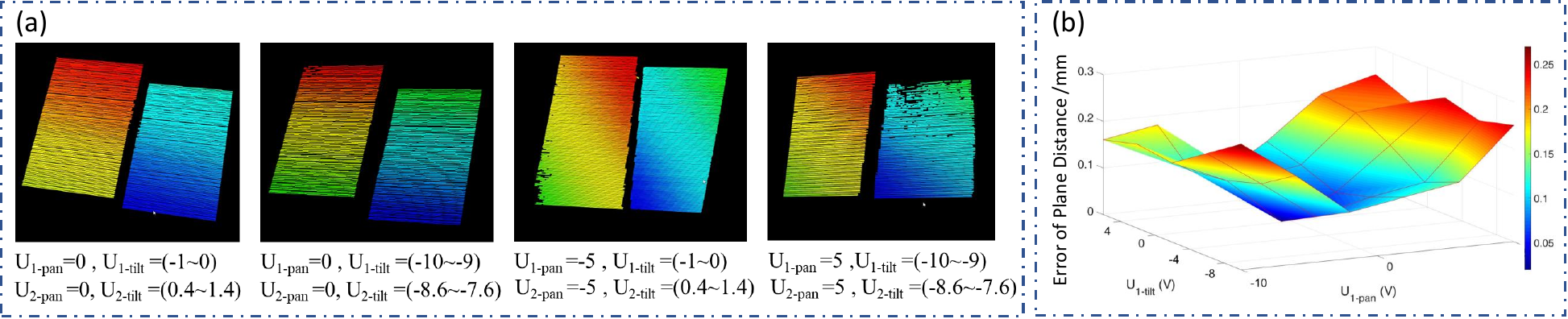}
	\caption{Standard blocks reconstruction test at different angles. (a) The point clouds of the stair at different angle. (b) 3D reconstruction error distribution at different angles.}
	\label{Fig12}
\end{figure*}

The reconstruction distance is 650 mm. The measurement range of the system is determined by the overlapping FOV of the dynamic camera and dynamic laser, which measures 1100 mm $\times$ 1300 mm. Thirty different positions are selected to analyze the reconstruction accuracy at different angles. The dynamic camera and laser simultaneously scan the target from these positions to complete the 3D reconstruction process. Fig. \ref{Fig12}(a) shows the example reconstructions obtained from four different positions, providing a visual representation of the reconstructed 3D models. The thickness error, which is related to the rotation angles of Galvanometer-1 and Galvanometer-2, is analyzed, as shown in Fig. \ref{Fig12}(b). It is evident from the graph that the error in 3D reconstruction increases as the rotation angles of the galvanometers deviate from their initial positions (calibration position). This is because larger errors occur as the system moves farther away from the calibration position. The RMSE for these thirty positions is calculated as 0.165 mm. These values provide evidence of the high precision achieved by the proposed system for 3D reconstruction.

 To compare the performance of the proposed method with that of existing methods \cite{article20, article21, article41, article44, article45, article42, article43}, we conduct comparative experiments using the standard component scanning method. The accuracy of the dynamic light-section 3D system depends primarily on the working distance. To perform a fair comparison, we repeat the standard component scanning procedure at various reconstruction distances, namely 100, 200, 350, 400, and 1000 mm, which are consistent with the working distances employed in existing methods.

\begin{figure}[H]
	\centering
	\includegraphics[width=0.37\textwidth]{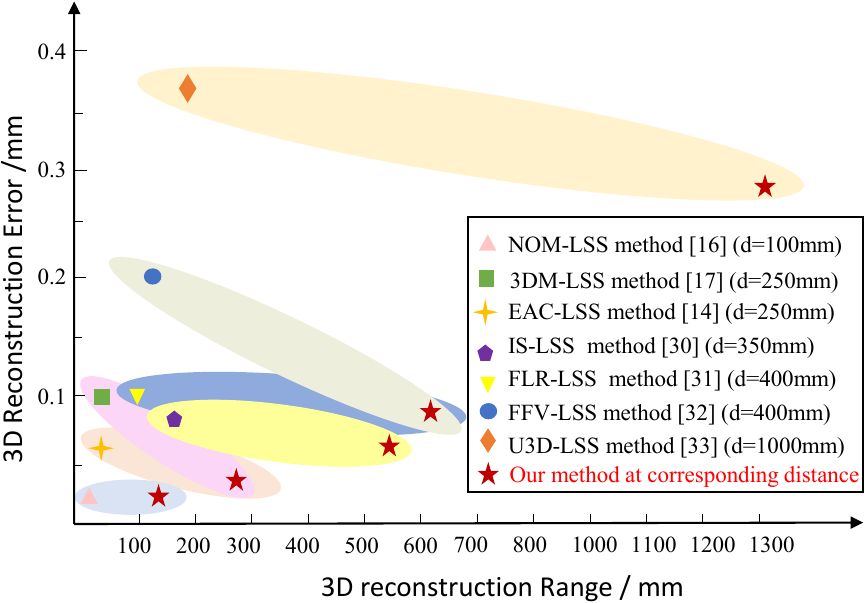}
	\caption{Comparison of 3D reconstruction accuracy and measurement ranges between existing method and proposed method.}
	\label{Fig13}
\end{figure}

The 3D reconstruction accuracy and measurement ranges achieved using each method are shown in Fig. \ref{Fig13}, with each color oval representing the same working distance. From the obtained results, it can be concluded that the proposed method exhibits smaller errors and larger measurement ranges than the existing methods at the corresponding working distances. This demonstrates the superior performance of our method in terms of accuracy and range compared with existing methods.

\subsubsection{Large object scanning test}
A high-precision machined large flat plate is also utilized to test the 3D reconstruction accuracy. The proposed 3D dynamic system is employed to scan the target and obtain its point clouds. The scanning distance is set at 650 mm. The measurement range is 1100 mm $\times$ 1300 mm. The size of the target is 740 mm $\times$ 740 mm. The position and angle of the target are changed arbitrarily within the depth of the field, and the reconstruction is repeated three times. After obtaining the reconstructed point cloud, a plane equation is fitted to the data using the RANSAC algorithm. 

The distances between all the points and the fitted plane are calculated, and the average value of these distances is considered as the error of the dynamic 3D system reconstruction. The RMSE for the three measurements is 0.281 mm. These results demonstrate that the proposed system achieves a high level of accuracy in 3D reconstruction measurements.

\section{Conclusion}
A dynamic light-section 3D reconstruction system is proposed in this study, which overcomes the trade-off between accuracy and measurement range by using multiple galvanometers. A mathematical model of the system is established, and a flexible and accurate calibration method is developed. The experimental results demonstrate that the proposed system performs well in terms of measurements, indicating its potential for industrial applications where high-precision and wide-range 3D reconstruction is required. Furthermore, the proposed method can be used in conjunction with the tracking algorithm for 3D reconstruction of moving targets.

\bibliographystyle{Bibliography/IEEEtranTIE}
\bibliography{Bibliography/Dynamic}\ %IEEEabrv instead of IEEEfull

\begin{IEEEbiography}[{\includegraphics[width=1in,height=1.25in,clip,keepaspectratio]{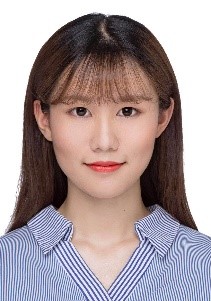}}]
{Mengjuan Chen} received the B.E. degree in Automation from Beijing University of Chemical Technology, Beijing, China, in 2016 and the master's degree in control engineering from the Institute of Automation, Chinese Academy of Sciences, Beijing, China. She is currently pursuing a Ph.D. degree at Hiroshima University, Hiroshima, Japan. Her research interests include high-speed image processing and 3D reconstruction.
\end{IEEEbiography}

\begin{IEEEbiography}[{\includegraphics[width=1in,height=1.25in,clip,keepaspectratio]{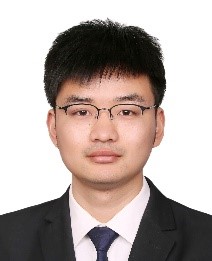}}]
	{Qing Li}  received the B.E. degree in Electronic and Information Engineering from the North University of China, China, in 2017. He received the M.E. degree in Information and Communication Engineering from the Beijing University of Technology, China, in 2020. is currently pursuing a PhD degree at Hiroshima University, Hiroshima, Japan. His research interest is high-speed image processing, mechanical control, and applications in industry and biomedicine.
\end{IEEEbiography}

\begin{IEEEbiography}[{\includegraphics[width=1in,height=1.25in,clip,keepaspectratio]{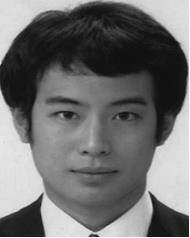}}]
	{Kohei Shimasaki} (Member, IEEE) received the B.E. degree in electrical, electronic, and systems engineering and the M.E. and Ph.D. degrees in system cybernetics from Hiroshima University, Higashi-Hiroshima, Japan, in 2016, 2018, and 2020, respectively. Since 2019, he has been a Designated Assistant Professor with the Digital Monozukuri (Manufacturing) Education Research Center, Hiroshima University. His current research interests include high-speed vision and vibration imaging.
\end{IEEEbiography}

\begin{IEEEbiography}[{\includegraphics[width=1in,height=1.25in,clip,keepaspectratio]{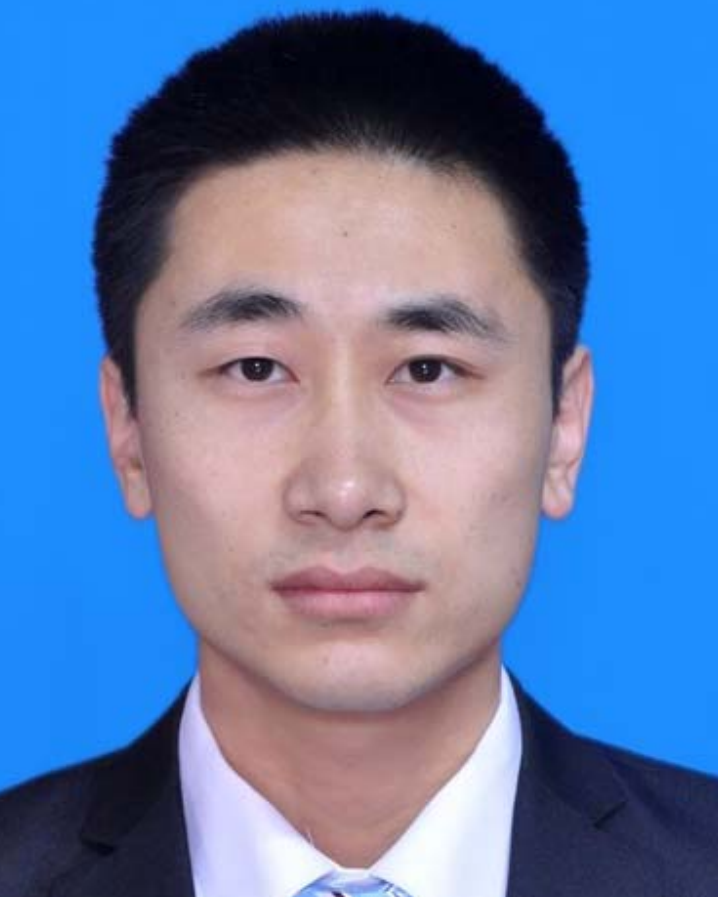}}]
	{Shaopeng Hu} received the B.E. and M.E. degrees from Northeastern University, China, in 2013 and 2015, respectively, and the Ph.D. degree from Hiroshima University, Japan, in 2018. From 2018 to 2020, he was a Designated Assistant Professor with Hiroshima University. He is currently an Assistant Professor with Hiroshima University. His research interests include high-speed vision, stereo measurement, and deep intelligence systems.
\end{IEEEbiography}

\begin{IEEEbiography}[{\includegraphics[width=1in,height=1.25in,clip,keepaspectratio]{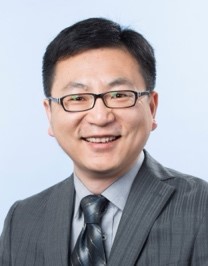}}]
{Qingyi Gu (M'13)} received the B.E. degree in Electronic and Information Engineering from Xi'an Jiaotong University, China, in 2005. He received the M.E. degree, and Ph.D. degree in Engineering, Hiroshima University, Japan, in 2010, and 2013 respectively. He is currently a professor at Institute of Automation, Chinese Academy of Sciences, China. His primary research interest is high-speed image processing and applications in industry and biomedicine.
\end{IEEEbiography}

\begin{IEEEbiography}[{\includegraphics[width=1in,height=1.25in,clip,keepaspectratio]{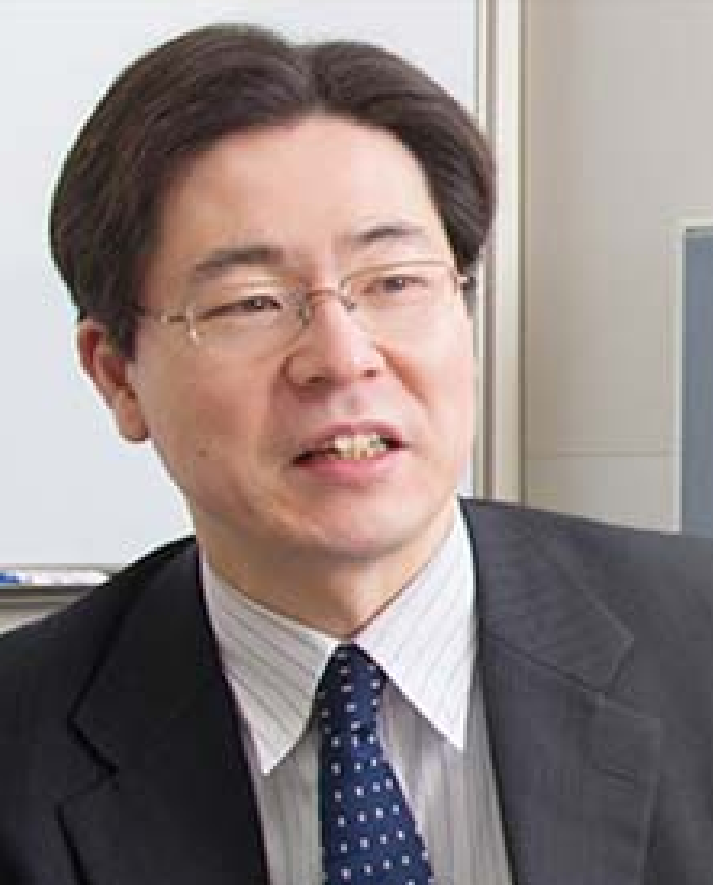}}]
	{Idaku Ishii (Member, IEEE)} received the B.E., M.E., and Ph.D. degrees from The University of Tokyo, Japan, in 1992, 1994, and 2000, respectively. He is currently a Professor at Hiroshima University, Japan. His research interests include high-speed robot vision, dynamic sensory information processing, sensor-based robot control, and their smart applications.
\end{IEEEbiography}

\end{document}